\documentclass{article}
\usepackage{arxiv}
\usepackage{cite}

\usepackage[
        colorlinks = true,
        linkcolor = blue,
        urlcolor  = blue,
        citecolor = blue,
]{hyperref}

\usepackage[utf8]{inputenc} 
\usepackage[T1]{fontenc}    
\usepackage{nicefrac}       
\usepackage{microtype}      
\usepackage{doi}
\usepackage{enumerate}
\usepackage{subfiles}
\usepackage[nottoc]{tocbibind}
\usepackage{lipsum}
\usepackage{mathtools}
\usepackage[noend]{algpseudocode}
\usepackage{enumitem,kantlipsum}
\usepackage{indentfirst}
\usepackage{multirow}
\usepackage{pifont}
\usepackage[table,xcdraw,dvipsnames,usenames]{xcolor}
\usepackage{subfloat}
\usepackage{amsmath,amssymb,amsfonts}
\usepackage{algorithm}
\usepackage{graphicx}
\usepackage{textcomp}
\usepackage{bbding}
\usepackage{rotating}
\usepackage{subcaption}
\usepackage{gensymb}
\usepackage{url}
\usepackage{booktabs}
\usepackage{breqn}

\usepackage[bottom,flushmargin]{footmisc}
\usepackage{adjustbox}
\usepackage{xcolor}
\usepackage{siunitx}

\usepackage{titlesec}
\titlespacing*{\section}{0pt}{0.5\baselineskip}{0.5\baselineskip}

\usepackage{makecell}

\hypersetup{
pdftitle={A template for the arxiv style},
pdfsubject={q-bio.NC, q-bio.QM},
pdfauthor={David S.~Hippocampus, Elias D.~Striatum},
pdfkeywords={First keyword, Second keyword, More},
}

\begin{document}
\begin{sloppypar}

\title{Lightweight Multi-System Multivariate Interconnection and Divergence Discovery}

\author{
    \href{https://orcid.org/0000-0003-2985-108X}{\includegraphics[scale=0.06]{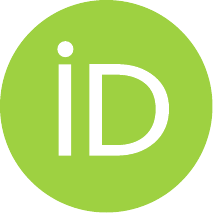}\hspace{1mm}Mulugeta~Weldezgina~Asres} \\
	Centre~for~Artificial~Intelligence~Research~(CAIR)\\
	University~of~Agder,~Norway\\
	\href{mulugetawa@uia.no}{mulugetawa@uia.no} \\
  \And
    \href{https://orcid.org/0000-0003-0299-171X}{\includegraphics[scale=0.06]{orcid.pdf}\hspace{1mm}Christian~Walter~Omlin} \\
	Centre~for~Artificial~Intelligence~Research~(CAIR)\\
	University~of~Agder,~Norway\\
	\href{christian.omlin@uia.no}{christian.omlin@uia.no} \\
 \And
     \href{https://orcid.org/0000-0002-1911-3158}{\includegraphics[scale=0.06]{orcid.pdf}\hspace{1mm}Jay~Dittmann} \\
	Baylor~University,~USA \\
    \href{jay_dittmann@baylor.edu}{jay_dittmann@baylor.edu} \\
\And
    \href{https://orcid.org/0000-0001-6743-3781}{\includegraphics[scale=0.06]{orcid.pdf}\hspace{1mm}Pavel~Parygin} \\
	University~of~Rochester,~USA \\
    \href{pavel.parygin@cern.ch}{pavel.parygin@cern.ch} \\
 \And
     \href{https://orcid.org/0000-0001-5921-5231}{\includegraphics[scale=0.06]{orcid.pdf}\hspace{1mm}Joshua~Hiltbrand} \\
	Baylor~University,~USA \\
    \href{joshua_hiltbrand@baylor.edu}{joshua_hiltbrand@baylor.edu} \\
 \And
     \href{https://orcid.org/0000-0002-4618-0313}{\includegraphics[scale=0.06]{orcid.pdf}\hspace{1mm}Seth~I.~Cooper} \\
	University~of~Alabama,~USA \\
    \href{seth.cooper@cern.ch}{seth.cooper@cern.ch} \\
 \And
     \href{https://orcid.org/0000-0002-8045-7806}{\includegraphics[scale=0.06]{orcid.pdf}\hspace{1mm}Grace~Cummings} \\
	University~of~Virginia,~USA \\
    \href{gec8mf@virginia.edu}{gec8mf@virginia.edu} \\
\And
     \href{https://orcid.org/0000-0001-5921-5231}{\includegraphics[scale=0.06]{orcid.pdf}\hspace{1mm}David~Yu} \\
	Brown~University,~USA \\
    \href{david_yu@brown.edu}{david_yu@brown.edu} \\
}

\maketitle

\begin{abstract}
Identifying outlier behavior among sensors and subsystems is essential for discovering faults and facilitating diagnostics in large systems. 
At the same time, exploring large systems with numerous multivariate data sets is challenging. 
This study presents a lightweight interconnection and divergence discovery mechanism (LIDD) to identify abnormal behavior in multi-system environments. The approach employs a multivariate analysis technique that first estimates the similarity heatmaps among the sensors for each system and then applies information retrieval algorithms to provide relevant multi-level interconnection and discrepancy details. 
Our experiment on the readout systems of the Hadron Calorimeter of the Compact Muon Solenoid (CMS) experiment at CERN demonstrates the effectiveness of the proposed method. Our approach clusters readout systems and their sensors consistent with the expected calorimeter interconnection configurations, while capturing unusual behavior in divergent clusters and estimating their root causes.

\end{abstract}


\section{Introduction}
\label{sec:introduction}

Multivariate analysis (MVA) is a statistical method for analyzing data involving multiple variables. It considers multiple variables to provide more accurate insights into the level of influence on variability and summarizes the relationships by preserving the important facets \cite{eide2018automated, abdi2010principal, mcdonald2014factor, hardoon2004canonical, bowen2011structural, borg2005modern,scholkopf1997kernel, van2008visualizing, mcinnes2018umap, haggag2014efficacy}. 
Various domains utilize MVA techniques for data reduction, grouping, clustering, dependency analysis, and hypothesis testing due to the multivariate nature of real-world problems \cite{eide2018automated}. 
Several methods have been employed for MVA, including but not limited to principal component analysis (PCA) \cite{abdi2010principal}, factor analysis (FA) \cite{mcdonald2014factor}, canonical correlation analysis (CCA) \cite{hardoon2004canonical}, structural equation modeling (SEM) \cite{bowen2011structural}, and multidimensional scaling (MDS) \cite{borg2005modern}.  
While these techniques capture linear relationships, there are also nonlinear methods such as non-linear manifold dimensionality reduction techniques, like kernel-PCA \cite{scholkopf1997kernel}, t-stochastic neighbor embedding (t-SNE) \cite{van2008visualizing}, and uniform manifold approximation and projection (UMAP) \cite{mcinnes2018umap}.
The best method to use generally depends on the type of data and the problem to be solved. 
This study focuses on an interpretable cluster analysis approach for discovering interconnections and discrepancies in multi-systems monitored with multivariate sensors.
Data clustering refers to automatically grouping similar objects based on measured data characteristics \cite{mullner2011modern}. 
Clustering approaches are adaptive to changes and enable outlier detection applications where divergent clusters are considered unusual behavior\cite{haggag2014efficacy, marfo2022condition}.

The \textit{Large Hadron Collider} (LHC) is the largest and highest-energy particle collider ever built \cite{evans2008lhc, cms2023development}. 
It is designed to conduct physics experiments that improve our understanding of the universe. 
The LHC is a two-ring superconducting accelerator and collider capable of accelerating and colliding beams of protons and heavy ions with the unprecedented luminosity of 10$^{34}$ cm$^{-2}$s$^{-1}$ and 10$^{27}$ cm$^{-2}$s$^{-1}$, respectively, 
at a center of mass energy of up to 13 TeV for proton-proton collisions and up to 5.36 TeV/nucleon for Pb-Pb collisions \cite{evans2008lhc, heuer2012future}. 
The CMS experiment, one of the two general-purpose detectors at the LHC, consists of two calorimeters: the \textit{electromagnetic} (ECAL) and the \textit{hadronic} (HCAL) to detect electrons, photons and hadrons, and a \textit{muon} system \cite{collaboration2008cms, cms2023development}.  
The HCAL consists of four subdetectors that utilize \textit{readout boxes} (RBXes) to incorporate frontend data acquisition electronics components \cite{collaboration2008cms, cms2023development}. 
The RBXes for a given subdetector share similar operations, technology, and configurations \cite{cms2023development}. 
Rapid identification and resolution of detector system abnormalities are required to yield high-quality particle data at the CMS experiment \cite{schneider2018data, pol2019anomaly, asres2021unsupervised, asres2022long, mulugeta2022dqm}. 
Thus, the electronic systems of the RBXes are monitored through several multivariate diagnostics sensor variables \cite{strobbe2017upgrade}. 
Most diagnostics sensors are employed in retrospective monitoring and debugging, which rely on simple statistical analysis and visual inspection of a large and diverse set of signals. There are also sensors used in online monitoring to generate alarms based on preset thresholds when values deviate from the expected range. 
Previous machine learning efforts on anomaly detection have extended RBX monitoring, capturing more challenging abnormalities, using pre-trained deep learning models on a set of sensor variables \cite{asres2021unsupervised, asres2022long, mulugeta2022dqm}.

Our study investigates online discrepancy discovery in the HCAL multi-RBX system configuration without prior training by analyzing deviations in sensor interconnection behavior. Our \textbf{lightweight interconnection and divergence discovery (LIDD)} \footnote{https://github.com/muleina/LIDD} aims to deliver a generic, scalable, and online fault detection and diagnostic system through multivariate analysis. 
Our experiment on the HCAL readout systems validates the effectiveness of the proposed method in clustering RBX systems and sensors consistent with the calorimeter's actual interconnection configurations. RBX systems with unusual sensor readings form divergent clusters, and our approach has identified the potential root cause of the divergence with visual illustrations and quantitative scores. 
Our approach shares similarities with CCA \cite{hardoon2004canonical} and MDS \cite{borg2005modern, hout2013multidimensional} as multivariate statistical analysis for the discovery and quantification of associations among sets of multivariate.
CCA runs on a pair of sets with multivariate, and MDS is a descriptive technique without statistical inference. 
The CCA and MDS, like most existing techniques, provide tools to display low data dimensions without delivering much interpretation \cite{van2008visualizing}.
In addition, the performance of these methods can be impacted by missing data, which is typical in real-world sensor measurements, as they expect tabular data. 
However, our approach tolerates measurement quality issues if the sensor interconnections within a given system are unaffected. 

In this paper, we describe the HCAL readout system in Section \ref{sec:background}. We briefly explain the monitoring sensor data sets in Section \ref{sec:datasetdescription}. We present our methods in Section \ref{sec:methodology}, and the evaluation and result discussion in Section \ref{sec:resultsanddiscussion}. We conclude the study in Section \ref{sec:conclusion}.

 \vspace*{-0.1\baselineskip}

\section{The Hadronic Calorimeter}
\label{sec:background}

\begin{figure}[!htbp]
\centering
\begin{subfigure}[]{0.6\columnwidth}
\centering
\includegraphics[width=1\columnwidth, scale=1]{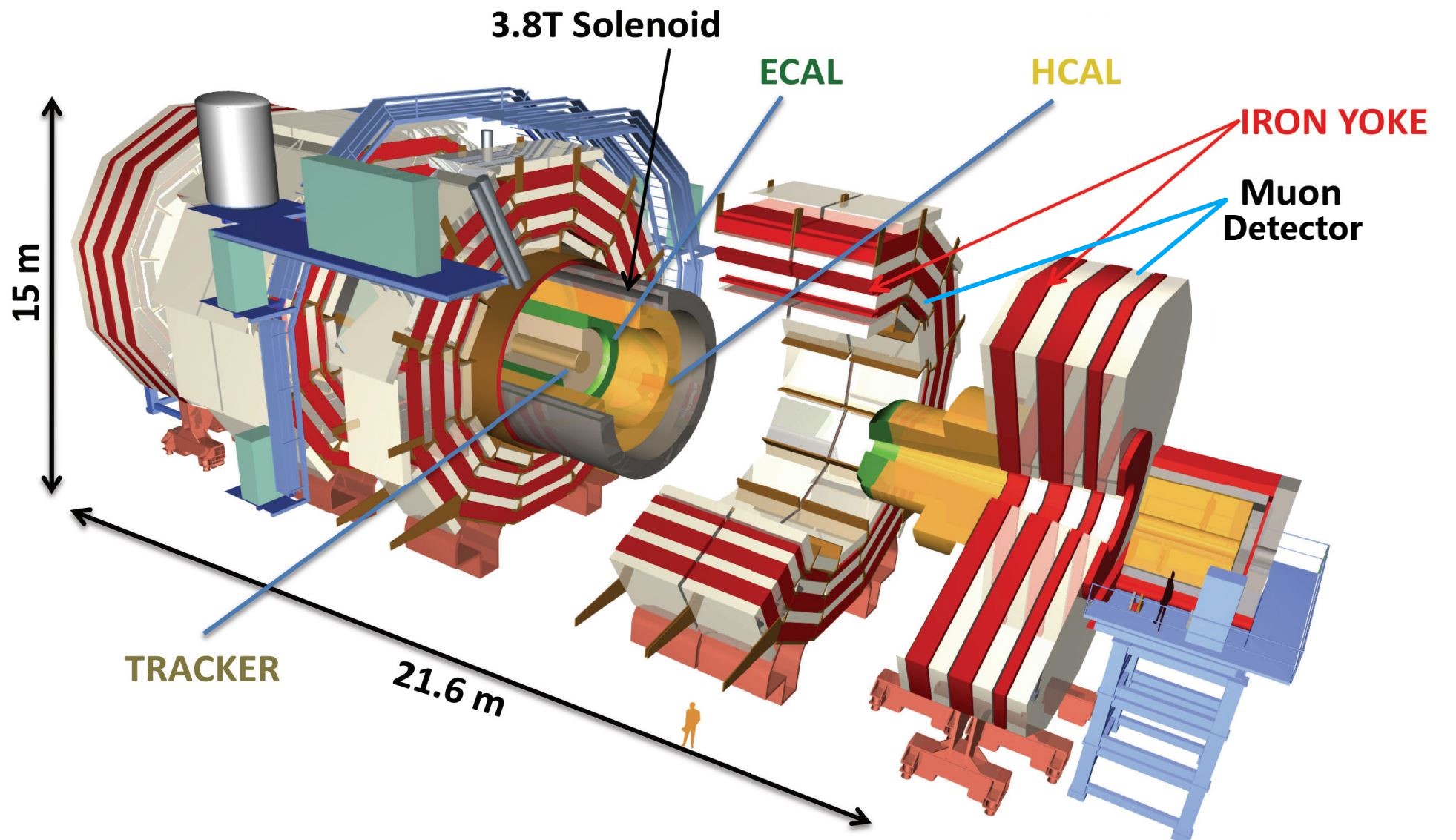}
\end{subfigure}
\caption{Schematic of the CMS detector with its major subsystems \cite{focardi2012status}.}
\label{fig:cms_diagram}
\end{figure}

\begin{figure}[!htbp]
\centering
\includegraphics[width=0.6\linewidth]{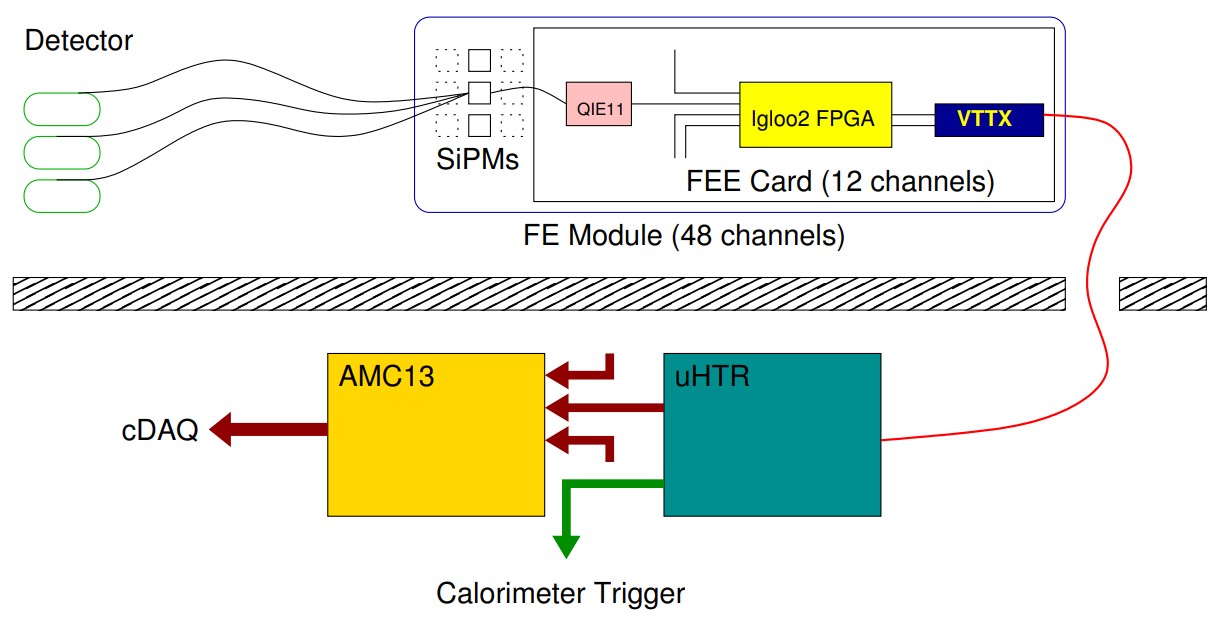}
\caption{The frontend data acquisition chain of the HE: the SiPMs, the QIE readout cards, and the optical link connected to the BE \cite{strobbe2017upgrade, cms2023development}. Each QIE card contains twelve QIE11 chips for charge integration, an Igloo2 field programmable gate array (FPGA) for data serialization and encoding, and a VTTx for optical transmission.}
\label{fig:HE_data_acquisition_system_chain}
 \vspace*{-\baselineskip}
\end{figure}

The HCAL is a specialized calorimeter that captures hadronic particles during a collision event in the CMS experiment (see Figure~\ref{fig:cms_diagram}) \cite{collaboration2008cms, mans2012cms, cms2023development}. 
The calorimeter is composed of brass and plastic scintillators, and the scintillation light produced in the plastic is transmitted via wavelength-shifting fibers to \textit{Silicon photomultipliers} (SiPMs) (see Figure~\ref{fig:HE_data_acquisition_system_chain}). 
The HCAL electronics systems are composed of two sections, known as the \textit{front-end electronics} (FE) and the \textit{back-end electronics} (BE) \cite{mans2012cms, cms2023development}.
The FE consist of components responsible for sensing and digitizing optical signals of the collision particles, and the BE receive data streams from the FE through fiber links and are composed of data preprocessing systems. 
The FE are divided into sectors of RBXes that house the electronics and provide voltage, backplane communications, and cooling.

The HCAL consists of four subdetectors: \textit{HCAL Endcap} (HE), \textit{HCAL Barrel} (HB), \textit{HCAL Forward} (HF), and \textit{HCAL Outer} (HO) \cite{mans2012cms, cms2023development}. The HE (the use-case of our study) uses 36 RBX sectors arranged on the plus (HEP) and minus hemispheres (HEM) of the CMS detector \cite{cms2023development}. Each RBX houses four \textit{readout modules} (RMs) for signal digitization \cite{strobbe2017upgrade, cms2023development}. Each RM has a SiPM control card, 48 SiPMs, and four readout \textit{charge integrating and encoding} (QIE) cards. 
The QIE card integrates charge from the SiPMs at 40 MHz, serializes and encodes the data, and finally, optically transmits to the backend system through the CERN \textit{versatile twin transmitter} (VTTx) at 4.8 Gbps.

\section{Dataset Description}
\label{sec:datasetdescription}
 
We utilized sensor data from the RM of the HE (HE-RM), and each RM has twelve diagnostic sensors: four from the SiPM control card and eight from the four readout QIE cards (see Table \ref{tbl:rca_interconn__rm_var_desc}). 
The dataset was obtained from the HCAL online software monitoring system (ngCCM server) from 01/08/2022 to 30/11/2022. 
The monitoring sensor data comprises four-month data of 20.7M samples, around 12K per sensor per RM. The data set is irregularly sampled and significantly sparse. The SiPM control card and QIE card sensors are logged about every eight and two hours, respectively. 
We downsampled the HE-RM-1 dataset, RM-1 data from all the 36 RBXes of the HE, to hourly intervals. We smoothed the data after cleaning short-lived transient spikes through a median filter and interpolating short gaps. Our main objective for the interconnection analysis is the time-persistent behavior of the RBXes.

\begin{table}[!h] 
\centering
\caption{HE-RBX-RM sensor data variables description}
\def\arraystretch{1.1}
\noindent
\resizebox{1.0\columnwidth}{!}{
\begin{tabular}{|l|c|c|c|}
\hline
\textbf{No.} & \textbf{Short Notation} & \textbf{Sensor Variable Name} & \textbf{Remark} \\ \hline
1            &      SPV             &     PELTIER\_VOLTAGE\_F                   &  Voltage reading of the SiPM Peltier temperature controller              \\ \hline
2            &     SPC              &      PELTIER\_CURRENT\_F                  &  Current reading of the SiPM Peltier temperature controller               \\ \hline
3            &     SRT              &      RTDTEMPERATURE\_F                  &  Temperature reading (averaged over 50 samples) by the SiPM control card               \\ \hline
4            &     SCH              &      HUMIDITY\_F                  &  Humidity reading by the SiPM control card               \\ \hline
5            &     Q1H              &      1-B-SHT\_RH\_F                  &  QIECARD 1 humidity reading             \\ \hline
6            &     Q2H              &      2-B-SHT\_RH\_F                  &  QIECARD 2 humidity reading             \\ \hline
7            &     Q3H              &      3-B-SHT\_RH\_F                  &  QIECARD 3 humidity reading             \\ \hline
8            &     Q4H              &      4-B-SHT\_RH\_F                  &  QIECARD 4 humidity reading             \\ \hline

9            &     Q1T              &      1-B-SHT\_TEMP\_F                  &  QIECARD 1 temperature reading             \\ \hline
10            &     Q2T              &      2-B-SHT\_TEMP\_F                  &  QIECARD 2 temperature reading             \\ \hline
11            &     Q3T              &      3-B-SHT\_TEMP\_F                  &  QIECARD 3 temperature reading             \\ \hline
12            &     Q4T              &      4-B-SHT\_TEMP\_F                  &  QIECARD 4 temperature reading             \\ \hline

\end{tabular}
}
\label{tbl:rca_interconn__rm_var_desc}
\end{table}
\section{Methodology}
\label{sec:methodology}

\begin{figure}[htbp]
\centering
\includegraphics[width=1\columnwidth]{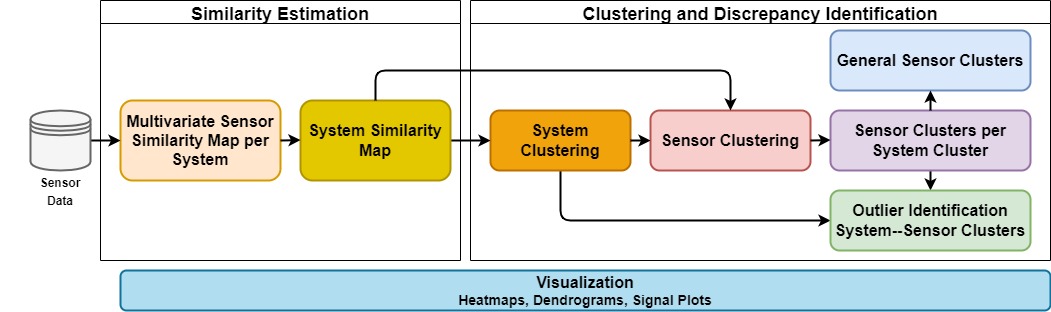} 
\caption{Our LIDD approach for systems and sensors interconnection and divergence discovery.}
\label{fig:rca__online_rca-interconnection_discovery}
\end{figure}

The interconnection discrepancy exploration approach in LIDD involves three main stages (see Figure~\ref{fig:rca__online_rca-interconnection_discovery}): 1) estimation of similarity among sensors per system and generates sensor interconnection maps, 2) estimation of similarity distance among the systems using their sensor interconnection maps, 3) system and sensor clustering analysis, and 4) divergence root-cause discovery.

\begin{itemize}

\item \textit{Sensor similarity map generation} ($I$): We estimate pairwise similarity $I^s$  among the multivariate sensors $X$ of a given system using $\Gamma$ function as:
\begin{equation}
\label{eq:rca__similarity_measure_sensor}
    I^s_k [i, j] = \Gamma (X_k [:, i], X_k [:, j]), ~i \neq j, ~ (i, j) \in S,  ~k \in M
\end{equation}
where $I^s_k \in \mathbb{R}^{N_s \times N_s}$ is the pairwise similarity matrix, and $X_k \in \mathbb{R}^{T \times N_s}$ is sensor data (with $N_s$ sensors and $T$ data samples) of the $k^{th}$ system in the system set $M$. The $\Gamma$ is a similarity measurement between the $i^{th}$ and $j^{th}$ sensors in the sensor set $S$. The $I^s$ is the similarity matrix map holding the pairwise score between the sensors. 
We employ a bivariate \textit{Pearson's correlation} ($\rho$) \cite{cohen2009pearson} for the $\Gamma$ for its fast computation and decent accuracy. The time delay effect is lessened in our use-case data, as the data sampling interval is fairly high. Pearson's correlation measures the linear correlation between two variables; it is the ratio between the covariance and the product of their standard deviations: 
\begin{equation}
    \rho({x_1}, {x_2}) = \frac{\sum_{i=1}^{T} ({x_1}[i] - {\bar{x}_1})({x_2}[i] - {\bar{x}_2})}{\sqrt{ \sum_{i=1}^{T} ({x_1}[i] - {\bar{x}_1})^2 }\sqrt{ \sum_{i=1}^{T} ({x_2}[i] - {\bar{x}_2})^2}}
\end{equation}
where $\rho$ is the correlation coefficient between sensor reading vector ${x_1}$ and ${x_2}$, and $\bar{x}_1$ and $\bar{x}_2$ denote the respective mean values of the vectors.
The $\rho \in [-1, 1]$ is a normalized covariance measurement; 
positive values for simultaneous increases and negative values for otherwise, small $| \rho |$ indicates a weak correlation, and zero implies no linear correlation. 
Other similarity methods like \textit{dynamic time wrapping} (DTW) and \textit{Granger's} may provide enhanced accuracy on time series data with higher computation overhead.

\item \textit{System similarity map estimation} ($D$): We utilize normalized pairwise Euclidean distance $\Phi$ between the sensor similarity maps of the systems to calculate similarity distance score matrix $D^m$ among the systems as:
\begin{equation}
\label{eq:rca__sys_distance_by_sensor_similarity_maps}
    D^m [k, h] = \Phi (I^s_k, I^s_h),  ~k \neq h,  ~(k, h) \in M
\end{equation}
where $D^m [k, h]$ is the similarity distance score between the $I^s$ of $k^{th}$ and $h^{th}$ systems. 
The $\Phi$ is calculated as:
\begin{equation}
  \Phi (I^s_k, I^s_h) = \frac{1}{N_s} \sqrt {\sum_{i\in S}^{N_s} \sum_{j\in S}^{N_s} ( I^s_k[i, j] - I^s_h[i, j] ) ^2 }
\end{equation}
where $\frac{1}{N_s}$ is the score normalizing factor.
\item \textit{System and sensor clustering analysis} ($C$ and $\xi$): 
We estimate the system and sensor clusters through hierarchical agglomerative clustering $\Theta$ with time complexity of $\mathcal{O}(n^2)$, where $n$ is the number of observations \cite{mullner2011modern}. 
The clustering link distance $C^m \in \mathbb{R}^{N_m \times N_m}$ of the systems is calculated on the $D$ using $\Phi$ and clustering optimization through nearest-neighbors chain algorithm \cite{mullner2011modern}:
\begin{equation}
\label{eq:rca__sys_clustering_scoring}
    C^m = \Theta (\Phi, D^m)
\end{equation}
The system clusters $\xi^m$ are generated by applying a threshold $\alpha^m$ on the $C^m$:
\begin{equation}
\label{eq:rca__sys_clustering_thr}
    L^m [k, h] = C ^m [k, h] < \alpha^m,~ k \neq h,  ~(k, h) \in M
\end{equation}
where $L^m \in \mathbb{Z}^{N_m \times N_m}$ holds the binary cluster links, with active edges for all the systems belonging to the same cluster. The $L^m$ generates $N_{\xi}$ system clusters, and the $i^{th}$ cluster $\xi^m_i$ is denoted as $\nu$, where $L^m [k, h]=1$ for $\{k, h\} \in \nu$.

We estimate clusters for the sensors using a multi-step approach: 1) we first estimate the representational sensor similarity map $I^m_\nu \in \mathbb{R}^{N_s \times N_s}$ per each system cluster $\nu$, and 2) we then generate sensor cluster links $L_\nu^s \in \mathbb{R}^{N_s \times N_s}$ from the $I^s_\nu$ for each $\nu$. 
Since each $\nu \in \xi^m$ represents distinct sensor interconnection characteristics, $I^s_\nu$ is calculated as the average sensor similarity map from all the systems in a cluster:
\begin{equation}
\label{eq:rca__sys_clustering_sensor_intercon}
   I^s_\nu [i, j] = \frac{1}{N_\nu} \sum_{k \in \nu}^{}{I^s_k [i, j]},  ~(i, j) \in S
\end{equation}
where $I^s_\nu [i, j]$ is the average similarity score between the $i^{th}$ and $j^{th}$ sensors for the $\nu$ system cluster. The $I^s_k$ and $N_\nu$ correspond to the sensor similarity map of the member system $k \in \nu$ and the number of systems in $\nu$, respectively.
We measure the pairwise sensor cluster distance score $C_\nu^s \in \mathbb{R}^{N_s \times N_s}$ per the given system cluster $\nu$ using $\Phi$ as:
\begin{equation}
\label{eq:rca__sensor_clustering_scoring_per_sys_cluster}
      C_\nu^s = \Theta (\Phi, I^s_\nu)
\end{equation}
The sensor clustering per $\nu$ are thus generated similarly by applying a threshold $\alpha^s$ on the $C_\nu^s$:
\begin{equation}
\label{eq:rca__sensor_clustering_thr_per_sys_cluster}
    L_\nu^s [i, j] = C_\nu^s [i, j] < \alpha^s,~ (i, j) \in S
\end{equation}

We estimate the overall sensor similarity map $I^s \in \mathbb{R}^{N_s \times N_s}$, the interconnection clustering distance $C^s \in \mathbb{R}^{N_s \times N_s}$, and cluster links $L^s \in \mathbb{Z}^{N_s \times N_s}$ by averaging over all the system clusters:
\begin{equation}
\label{eq:rca__sensor_clustering}
\begin{aligned}
 & I^s [i, j] = \frac{1}{N_{\xi}} \sum_{\nu \in \xi^m}^{}{I^s_\nu [i, j]} \\
 & C^s = \Theta (\Phi, I^s) \\
& L^s [i, j] = C^s [i, j] < \alpha^s,~ i \neq j,  ~(i, j) \in S
\end{aligned}
\end{equation}

\item \textit{Divergence root-cause discovery} ($\psi$ and $R$): We locate the root-cause sensor variables that contribute significantly to cluster split among the systems using the aggregated difference of the average interconnection maps $I^s_\nu$ of the system clusters as:
\begin{equation}
\label{eq:rca__rca_scoring}
    \psi^s_\nu [\upsilon, i] = \Phi (I^s_\nu[i, :], I^s_\upsilon[i, :]), ~(\nu, \upsilon) \in \xi^m,~i \in S
\end{equation}
where $\psi^s_\nu [\upsilon, i]$ is the sensor interconnection divergence score of the system cluster $\nu$ from cluster $\upsilon$ at the $i^{th}$ sensor variable. The overall sensor divergence score $\bar{\psi}^s_\nu \in \mathbb{R}^{1 \times N_s}$ for a given system cluster $\nu$ is estimated as: 
\begin{equation}
\label{eq:rca__rca_scoring_agg}
    \bar{\psi}^s_\nu [i] = \sum_{\upsilon \in \xi^m}^{}{\psi^s_\nu [\upsilon, i],~i \in S}
\end{equation} 
We apply a threshold $\alpha^\phi$ to select the significant root cause contributions $R^s_\nu$ as:
\begin{equation}
\label{eq:rca__rca_flag}
    R^s_\nu [i] = \bar{\psi}^s_\nu > \alpha^\phi
\end{equation}
\end{itemize}

\section{Results and Discussion}
\label{sec:resultsanddiscussion}

This section presents the results of our proposed LIDD approach in a multi-system multivariate interconnection discrepancy experiment. We apply the approach to the 36 RBX-RM systems of the HE, each monitored by twelve sensor variables. 

\subsection{System Interconnections Discovery}

\begin{figure}[!h]
\centering
\includegraphics[width=0.8\columnwidth]{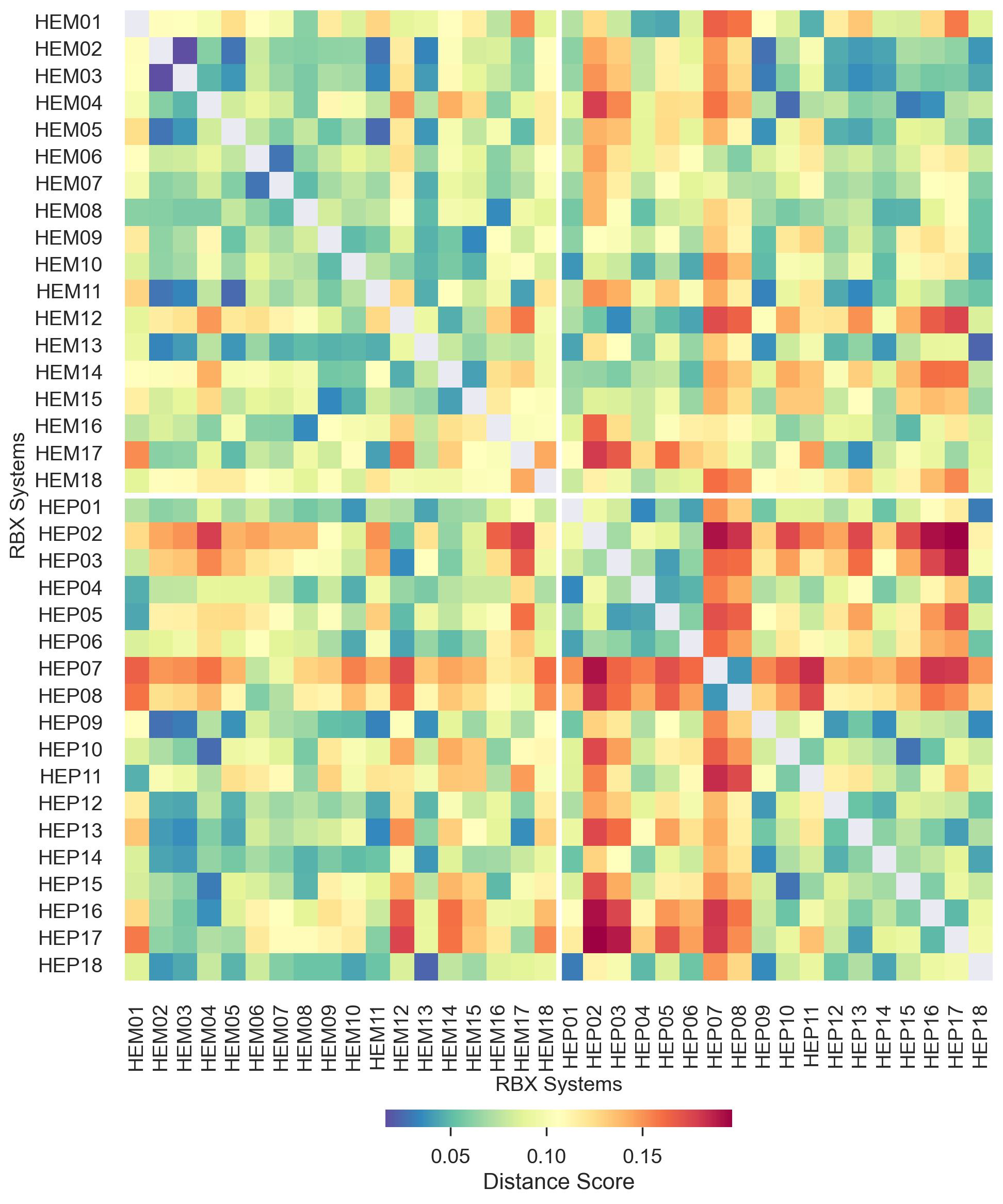}
\caption{RBX multi-system pairwise dissimilarity distance heatmap. The color bar shows the dissimilarity score $D$ (the normalized Euclidean distance on sensor interconnection maps).}
\label{fig:rca__multi_sensors_corrmap_distance_among_rbxes_heatmap}
 \vspace*{-\baselineskip}
\end{figure}

\begin{figure}[!h]
\centering
\includegraphics[width=0.75\columnwidth]{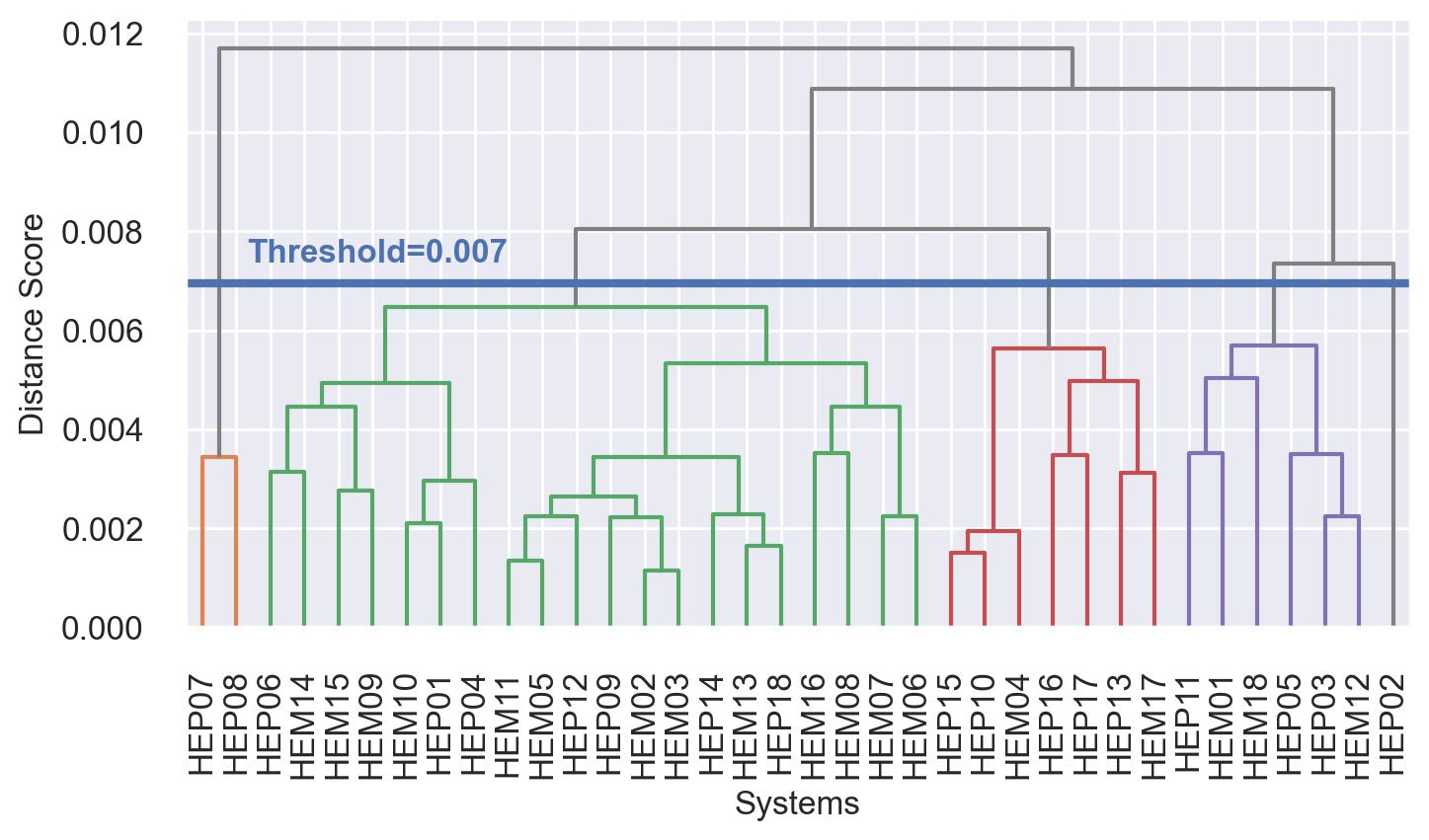}
\caption{RBX system clustering based on similarity on multivariate sensor interconnection at the distance threshold $\alpha^m=0.007$. Five clusters ($N_{\xi}$=5) are generated: \textcolor{orange}{CL-1}, \textcolor{green}{CL-2}, \textcolor{red}{CL-3}, \textcolor{violet}{CL-4}, and \textcolor{gray}{CL-5}, where the CL-i denotes the $i^{th}$ cluster. }
\label{fig:rca__multi_sensors_corrmap_distance_among_rbxes_clustered_dendrogram}
 \vspace*{-\baselineskip}
\end{figure}

\begin{figure}[!htbp]
\centering
\begin{subfigure}[]{0.33\columnwidth}
\centering
\includegraphics[width=\linewidth]{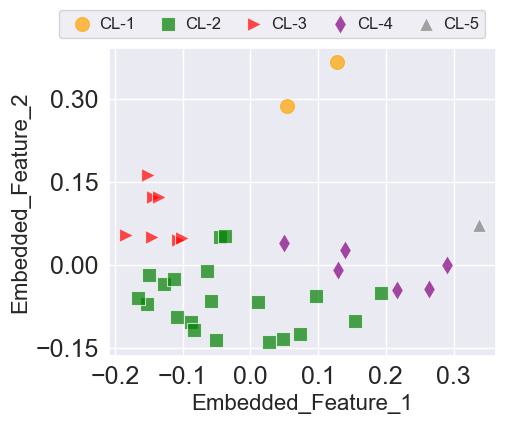}
\end{subfigure}
\begin{subfigure}[]{0.33\columnwidth}
\centering
\includegraphics[width=\linewidth]{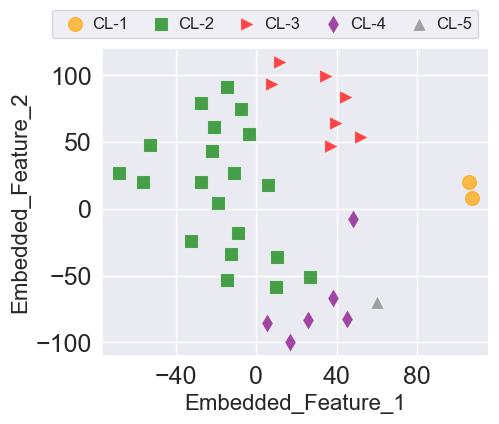}
\end{subfigure}
\begin{subfigure}[]{0.316\columnwidth}
\centering
\includegraphics[width=\linewidth]{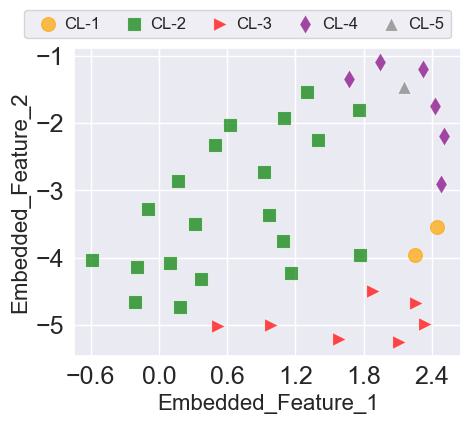}
\end{subfigure}
\caption{Dimension reduction (36 $\rightarrow 2$ embedded feature vectors) on the similarity distance score $D$ using (top-left) PCA \cite{abdi2010principal}, (top-right) t-SNE \cite{van2008visualizing}, and (bottom-center) UMAP \cite{mcinnes2018umap}. The overlaps among the clusters suggest that a higher number of features for dimension reduction is required.}
\label{fig:rca__multi_sensors_corrmap_distance_among_rbxes_clustered_view_umap}
 \vspace*{-1.\baselineskip}
\end{figure}

The heatmap plot in Figure~\ref{fig:rca__multi_sensors_corrmap_distance_among_rbxes_heatmap} and clustering diagram in Figure~\ref{fig:rca__multi_sensors_corrmap_distance_among_rbxes_clustered_dendrogram} illustrate the behavioral dissimilarity among the RBX systems, respectively. The sensor pairwise correlation maps of the RBXes determine the distance, and thus, the RBXes are clustered based on their similarity in sensor interconnection behavior. 
The outlying clusters can be easily detected, considering their size and distance from the rest of the clusters. 
The RBX clustering indicates outlier behavior at \textcolor{orange}{cluster-1 (CL-1)}: \{HEP07, HEP08\}, and \textcolor{gray}{cluster-5 (CL-5)}: \{HEP02\} with a small number of cluster members and significant distance as shown in the dendrogram. 
For comparison, we have applied dimension reduction techniques, such as PCA \cite{abdi2010principal}, t-SNE \cite{van2008visualizing}, and UMAP \cite{mcinnes2018umap}, to the system dissimilarity matrix $D$ to visualize the clusters in two-dimensional (see Figure~\ref{fig:rca__multi_sensors_corrmap_distance_among_rbxes_clustered_view_umap}). The figures demonstrate that reduction algorithms can capture the variations but may struggle to provide a clear split in some clusters. Furthermore, identifying the divergence root causes from the dimension reduction algorithms is not straightforward and remains challenging. 
Next, we will discuss our proposed mechanisms for discovering diagnostic root causes of the estimated RBX system clusters.

\subsection{Sensor Interconnections Discovery}

\begin{figure}[!htbp]
\centering

\begin{subfigure}[]{\columnwidth}
\centering
\begin{subfigure}[]{0.32\columnwidth}
\centering
\includegraphics[width=1\linewidth]{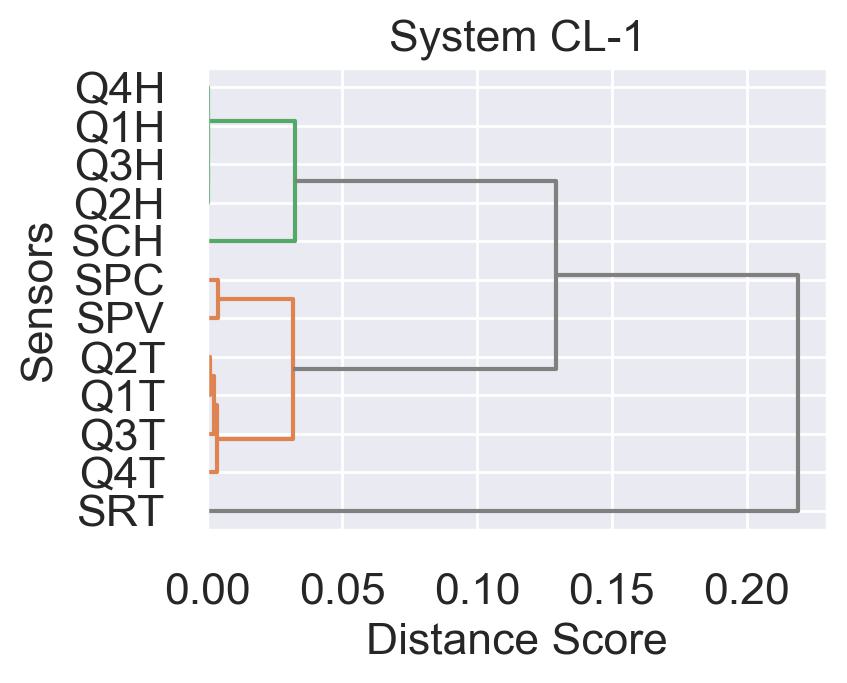}
\end{subfigure}
\begin{subfigure}[]{0.32\columnwidth}
\centering
\includegraphics[width=1\linewidth]{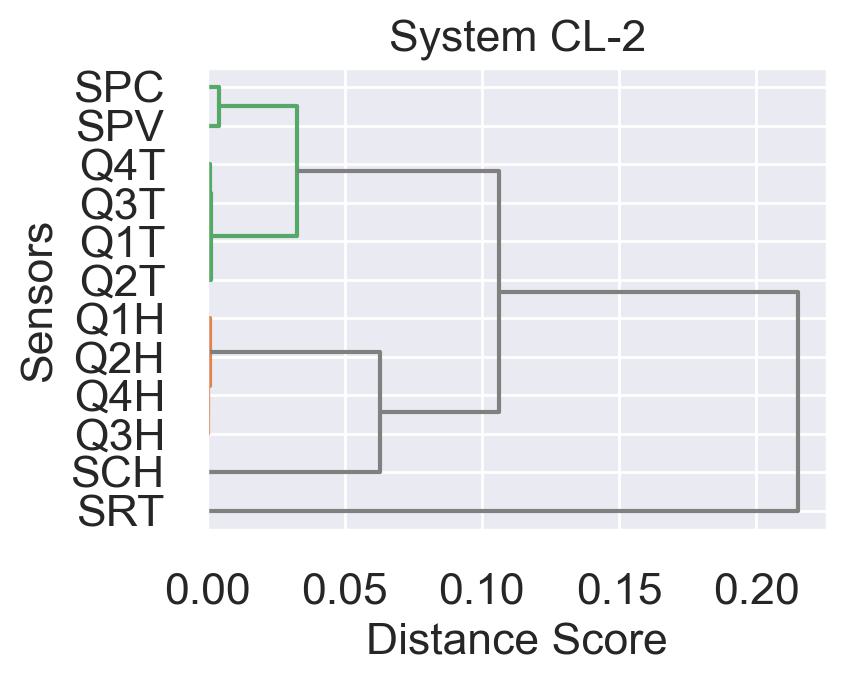}
\end{subfigure}
\begin{subfigure}[]{0.32\columnwidth}
\centering
\includegraphics[width=1\linewidth]{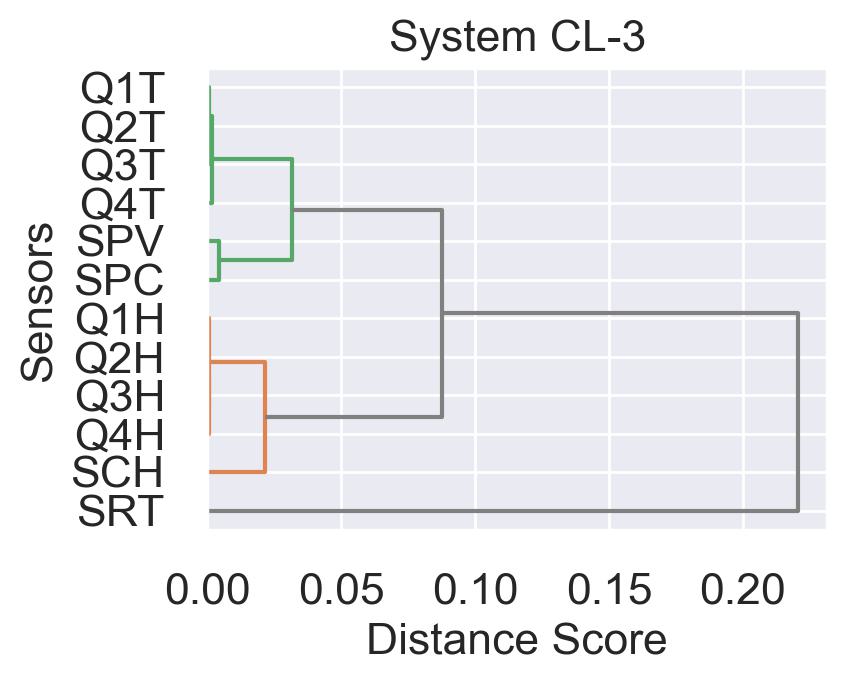}
\end{subfigure}
\end{subfigure}
\begin{subfigure}[]{\columnwidth}
\centering
\begin{subfigure}[]{0.32\columnwidth}
\centering
\includegraphics[width=1\linewidth]{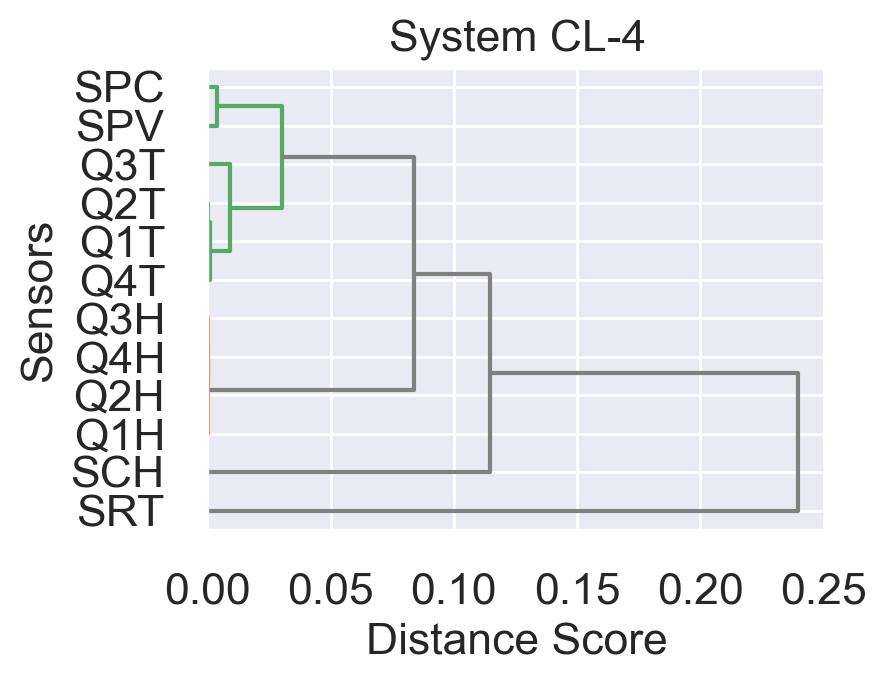}
\end{subfigure}
\begin{subfigure}[]{0.32\columnwidth}
\centering
\includegraphics[width=1\linewidth]{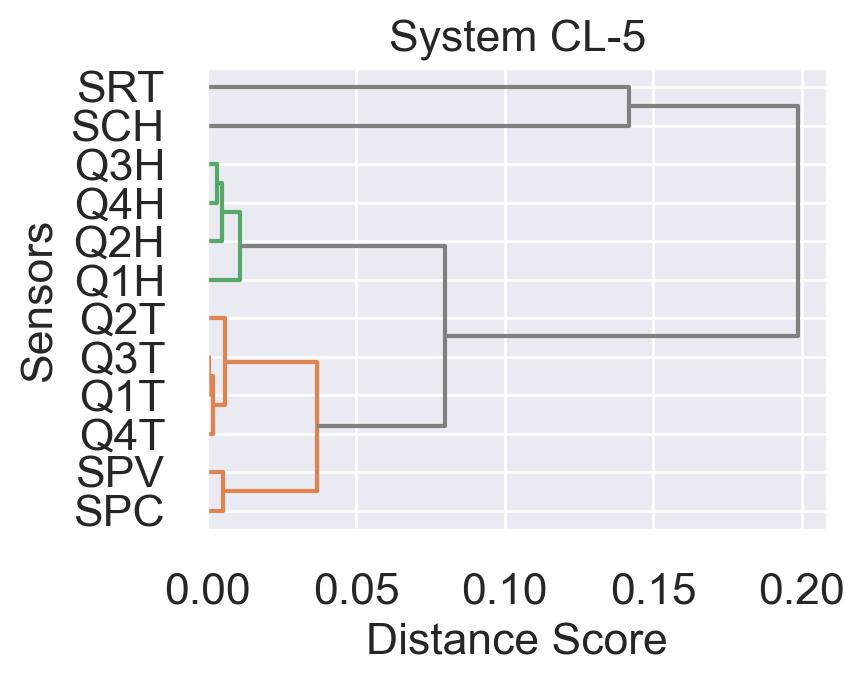}
\end{subfigure}
\begin{subfigure}[]{0.32\columnwidth}
\centering
\includegraphics[width=1\linewidth]{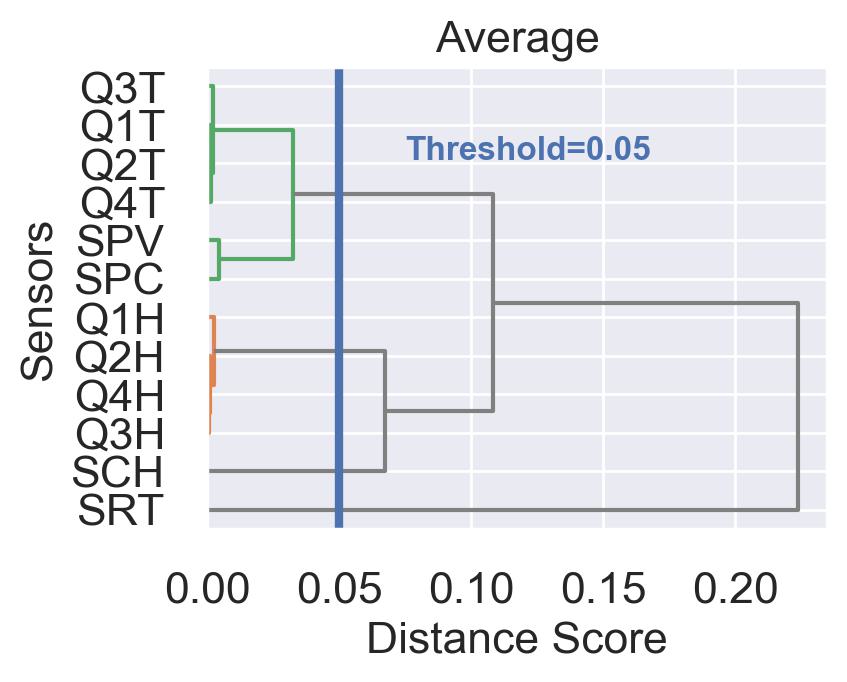}
\end{subfigure}
\end{subfigure}

\caption{Multivariate sensor interconnection clustering dendrogram per RBX system cluster using sensor clustering threshold $\alpha^s=0.05$. 
The average over the system clusters indicates substantial discrepancies in the SRT and SCH sensors, and the interconnection strength.
}
\label{fig:rca__sensors_clustered_dendrogram_per_rbx_clusters}
 \vspace*{-\baselineskip}
\end{figure}

\begin{figure*}[!htbp]
\centering
\includegraphics[width=1\columnwidth]{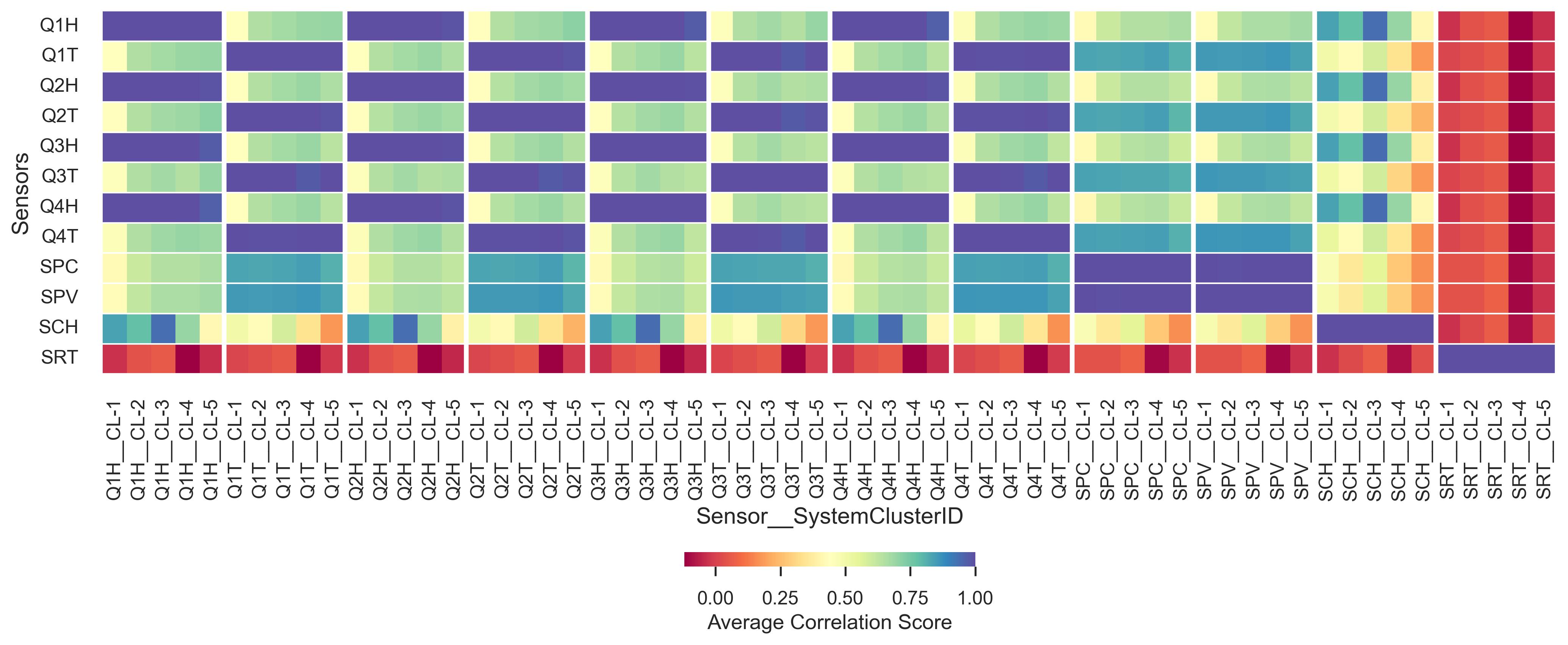}
\caption{
The heatmap of the RBX system clusters interconnection on the multivariate sensors. The group boxes show the interconnection strength of the sensors on the $x-axis$ for system clusters on the $y-axis$. The divergent colors within each box indicate outlier characteristics, e.g., SCH, Q[1-4]H, and Q[1-T] sensors.
}
\label{fig:rca__multi_sensors_corrmap_distance_among_rbx_clusters_vs_sensors_heatmap_clthr}
 \vspace*{-\baselineskip}
\end{figure*}

\begin{figure*}[!htbp]
\centering
\includegraphics[width=1\textwidth]{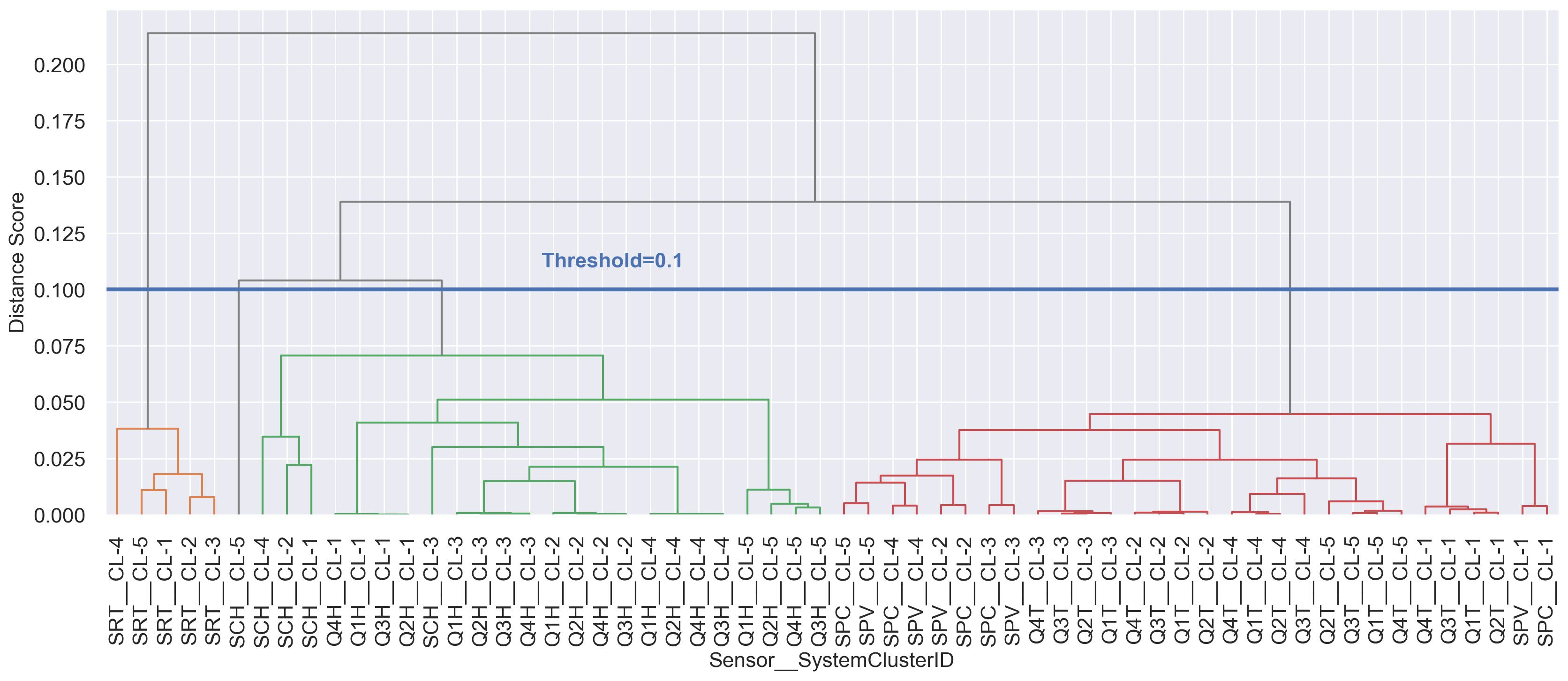}
\label{fig:rca__multi_sensors_corrmap_distance_among_sensors_clustered_dendrogram_clthr_higher}
\caption{Dendrogram of sensor interconnections across RBX system clusters at $\alpha^s=0.1$. 
The sensor clustering can be adjusted by increasing and decreasing the $\alpha^s$ to capture solid and subtle differences, respectively, among the RBX clusters. For instance, lowering the threshold to $\alpha^s=0.05$ will isolate the Q[1-4]H of the CL-5, and the SCH from the Q[1-4]H sensors.
}
\label{fig:rca__multi_sensors_corrmap_distance_among_sensors_clustered_dendrogram_clthr}
 \vspace*{-\baselineskip}
\end{figure*}

The representative sensor interconnections for the estimated system clusters, calculated from Eqs. \eqref{eq:rca__sensor_clustering_scoring_per_sys_cluster} and \eqref{eq:rca__sensor_clustering_thr_per_sys_cluster}, are portrayed in Figure~\ref{fig:rca__sensors_clustered_dendrogram_per_rbx_clusters}--\ref{fig:rca__multi_sensors_corrmap_distance_among_sensors_clustered_dendrogram_clthr}. 
Figure~\ref{fig:rca__sensors_clustered_dendrogram_per_rbx_clusters} presents the sensor interconnections clustered dendrograms for a given RBX cluster ($L^s_\nu$) estimated using \eqref{eq:rca__sensor_clustering_thr_per_sys_cluster}. 
The dendrograms highlight differences in the interconnection of SCH and SRT sensors among RBX clusters.
Figure~\ref{fig:rca__sensors_clustered_dendrogram_per_rbx_clusters} (bottom-right) depicts the average sensor interconnection aggregating the distinct characteristics across the RBX clusters using \eqref{eq:rca__sensor_clustering}. 
Figure~\ref{fig:rca__multi_sensors_corrmap_distance_among_rbx_clusters_vs_sensors_heatmap_clthr} provides the visual diagnostics heatmap through $I^s_\nu$s to detect any discrepancies in the sensor connections across the RBX system clusters. 
The difference in coloring within each column box indicates which sensors in the RBX clusters behave differently, while uniform coloring indicates behavior similarity. For example, the SCH and Q[1-4]H sensors show discrepancies among the clusters.
Figure~\ref{fig:rca__multi_sensors_corrmap_distance_among_sensors_clustered_dendrogram_clthr} depicts the equivalent clustering dendrogram of Figure~\ref{fig:rca__multi_sensors_corrmap_distance_among_rbx_clusters_vs_sensors_heatmap_clthr} to capture the differences quickly besides the heatmap visualization.
The figure portrays the clustered multivariate sensor interconnections across all the RBX system clusters; sensors measuring similar quantities across similar subsystems are clustered together---e.g., SiPM card: SPV, SPC, SRT, and SCH, and QIE cards: Q[1-4]T and Q[1-4]H. The behavioral similarity among the four QIE card sensors within an RBX cluster is more substantial than across RBX clusters. The SCH sensor of the RBX CL-5 is divergent. 

The generated general knowledge of the system and sensor interconnection from the above clustering illustrations can be summarized as: 1) the SPV and SPC are strongly interconnected, 2) the Q[1-4]T are strongly interrelated and connected to the SPV and SPC, 3) the Q[1-4]H are strongly interrelated and connected to the SCH but with weaker strength, and 4) the SRT sensor has distinct and isolated behavior.
The association among the Q[1-4]H and Q[1-4]T weakens (distance increases) at the RBX CL-4. The difference in the sampling method used for the sensors might be responsible for the SRT; the SRT reading is an average of 50 samples, whereas the other sensors record immediate values at 1-minute sampling intervals.

\subsection{Divergence Root-Cause Discovery}

\begin{figure}[!htbp]
\centering

\begin{subfigure}[]{\columnwidth}
\centering
\begin{subfigure}[]{0.37\linewidth}
\centering
\includegraphics[width=0.8\linewidth]{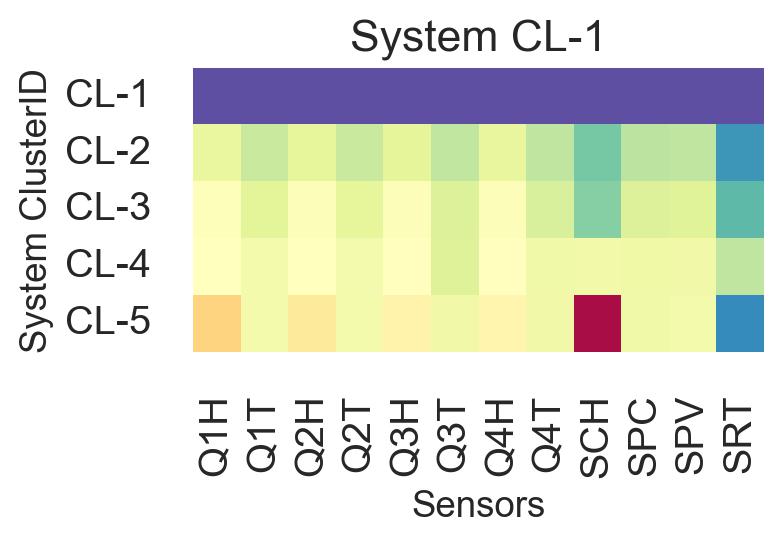}
\end{subfigure}
\begin{subfigure}[]{0.3\linewidth}
\centering
\includegraphics[width=0.8\linewidth]{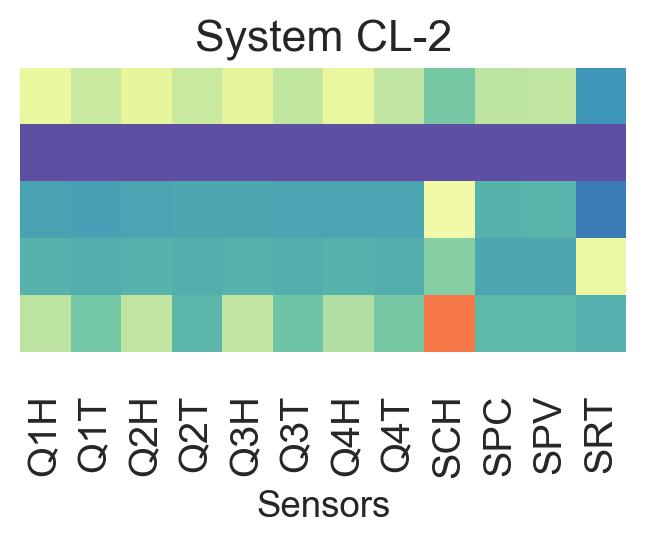}
\end{subfigure}
\begin{subfigure}[]{0.3\linewidth}
\centering
\includegraphics[width=0.8\linewidth]{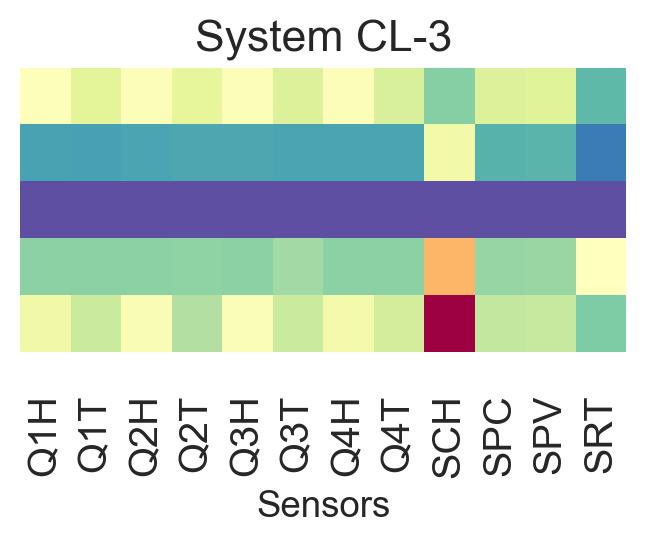}
\end{subfigure}
\end{subfigure}

\begin{subfigure}[]{\columnwidth}
\centering
\begin{subfigure}[]{0.38\linewidth}
\centering
\includegraphics[width=0.8\linewidth]{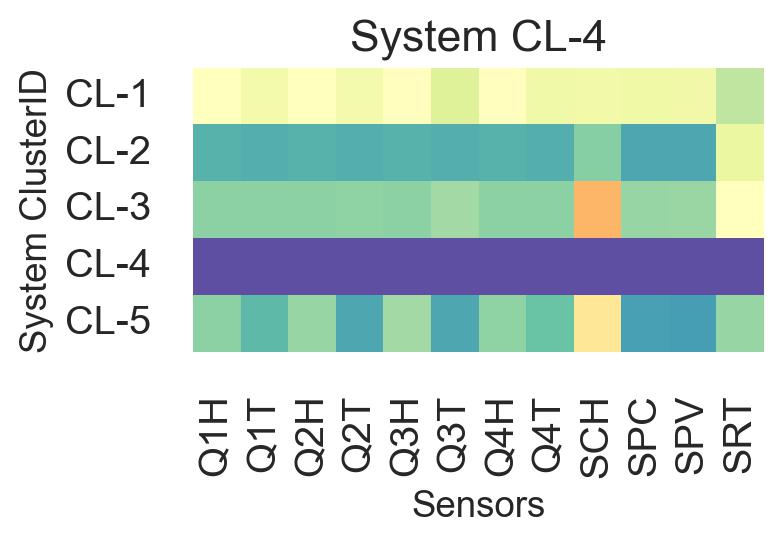}
\end{subfigure}
\begin{subfigure}[]{0.38\linewidth}
\centering
\includegraphics[width=0.8\linewidth]{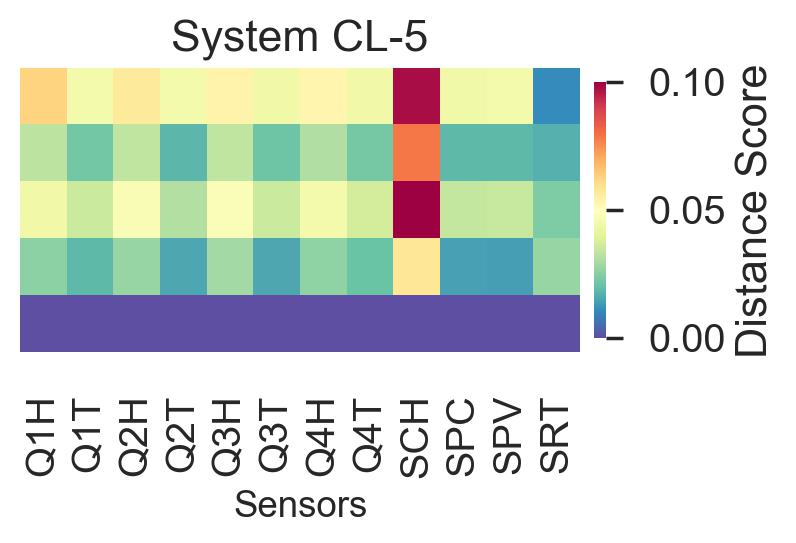}
\end{subfigure}
\caption{}
\end{subfigure}

\begin{subfigure}[]{\columnwidth}
\centering
\begin{subfigure}[]{0.48\linewidth}
\centering
\includegraphics[width=0.8\linewidth]{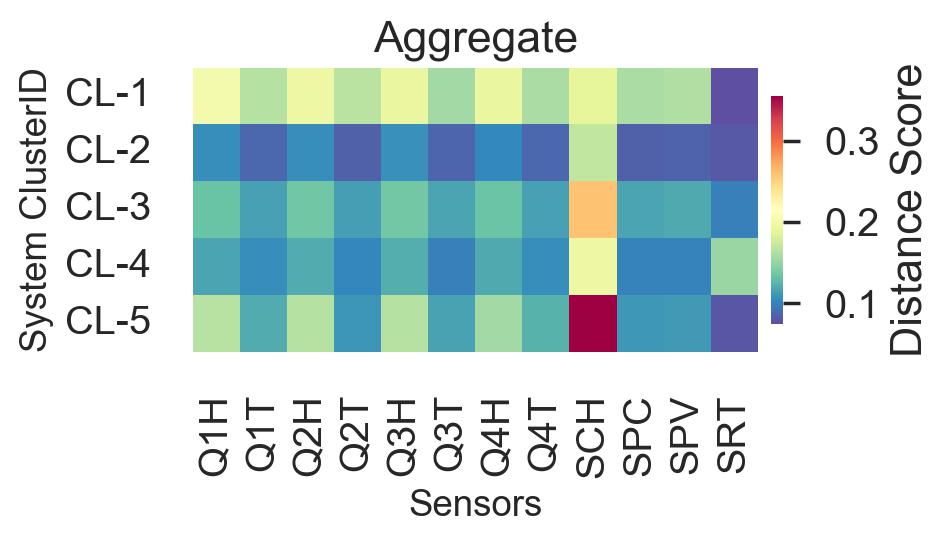}
\caption{}
\label{fig:rca__multi_sensors_among_rbx_clusters_diff_summary_heatmap}
\end{subfigure}
\begin{subfigure}[]{0.46\linewidth}
\centering
\includegraphics[width=0.8\linewidth]{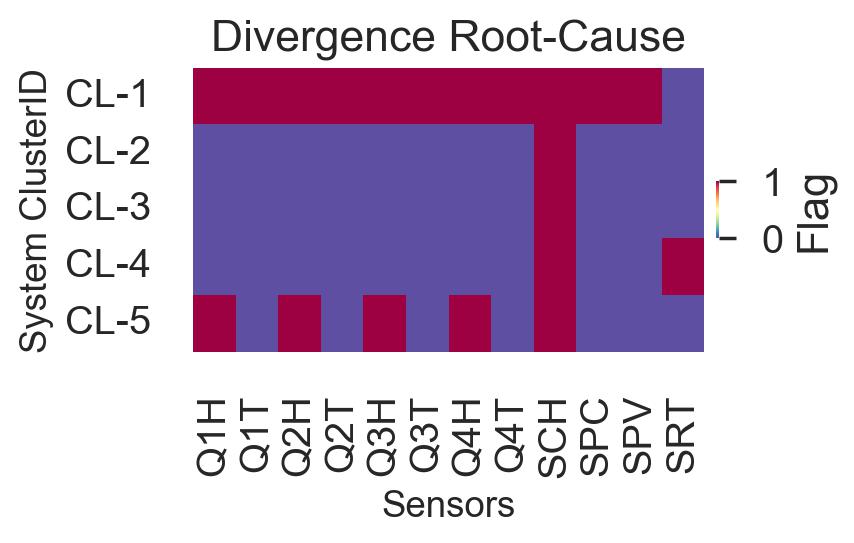}
\caption{}
\label{fig:rca__multi_sensors_among_rbx_clusters_diff_summary_heatmap_thr}
\end{subfigure}
\end{subfigure}

\caption{Divergence root-cause detection using the difference in sensor interconnections among the RBX systems clusters. Plots a) illustrate the sensor divergence score $\psi^s_\nu$ of each system cluster, and the color bars show the strength of discrepancy. The plots in b) and c) are the aggregate divergence scores $\bar{\psi}^s_\nu$ and the root-cause flags after threshold $\alpha^\phi=0.15$, respectively. The plots indicate the noticeable divergence in the SCH sensors in all clusters, Q[1-4]H in CL-1 and CL-5, and SRT in CL-4.
}
\label{fig:rca__multi_sensors_among_rbx_clusters_diff}
 \vspace*{-\baselineskip}
\end{figure}

We capture the diagnostics discrepancies quantitatively, in addition to the visual representation, to identify the causes of the clustering divergences using Eqs. \eqref{eq:rca__rca_scoring}--\eqref{eq:rca__rca_flag}. Figure~\ref{fig:rca__multi_sensors_among_rbx_clusters_diff} depicts the score of the discrepancies heatmaps among the RBX clusters for each sensor variable. The plots show where the difference lies for each system cluster compared to the others.  
Figure~\ref{fig:rca__multi_sensors_among_rbx_clusters_diff_summary_heatmap} and  \ref{fig:rca__multi_sensors_among_rbx_clusters_diff_summary_heatmap_thr} provide the aggregate summary score $\bar{\psi}^s_\nu$ and the generated root-cause detection flags $R^s_\nu$, respectively. The illustrations indicate that the SCH sensor is the main contributor in most clusters, and most clustering CL-1 sensors exhibit discrepancies. The results validate that the proposed approach successfully captured the summary root causes visually illustrated in the interconnection heatmaps and clustering dendrograms. The summary of root causes is essential for facilitating diagnostics, particularly when the number of clusters or sensors is large, and examining extensive heatmaps or dendrograms might be challenging. 

\begin{figure*}[!htbp]
\centering
\begin{subfigure}[]{1.00\textwidth}
\centering
\includegraphics[width=\linewidth]{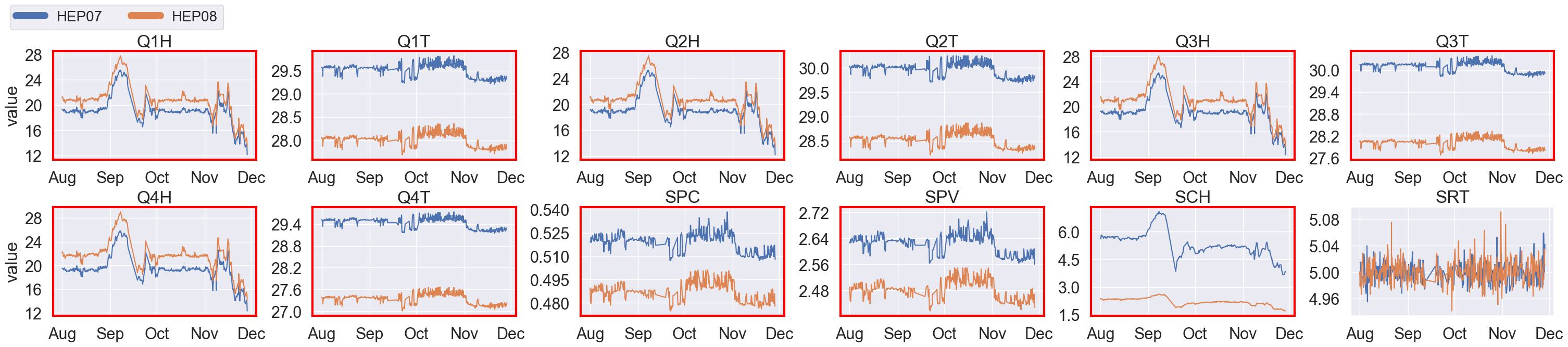}
\end{subfigure}

\begin{subfigure}[]{1.00\textwidth}
\centering
\includegraphics[width=\linewidth]{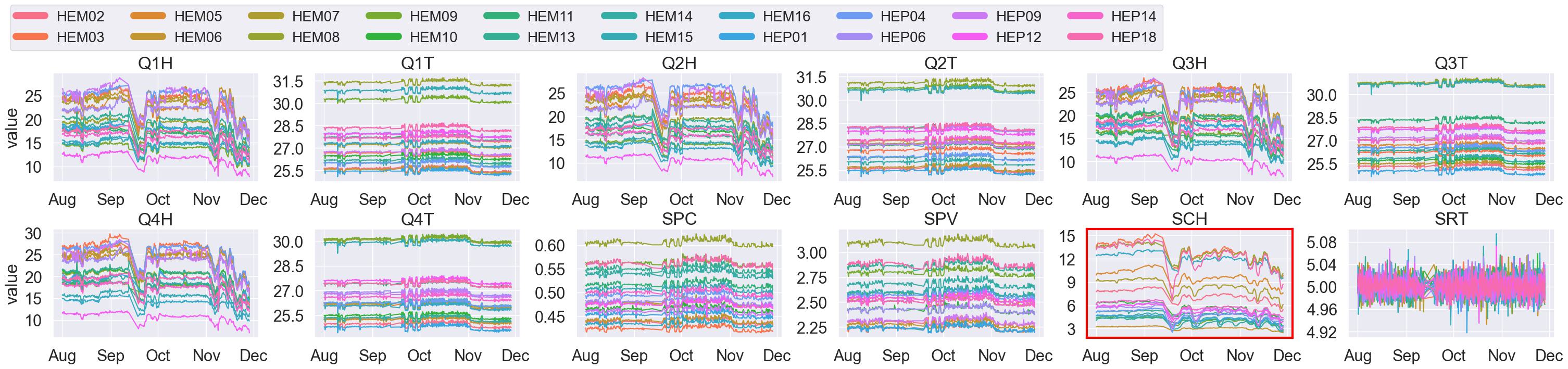}
\end{subfigure}

\begin{subfigure}[]{1.00\textwidth}
\centering
\includegraphics[width=\linewidth]{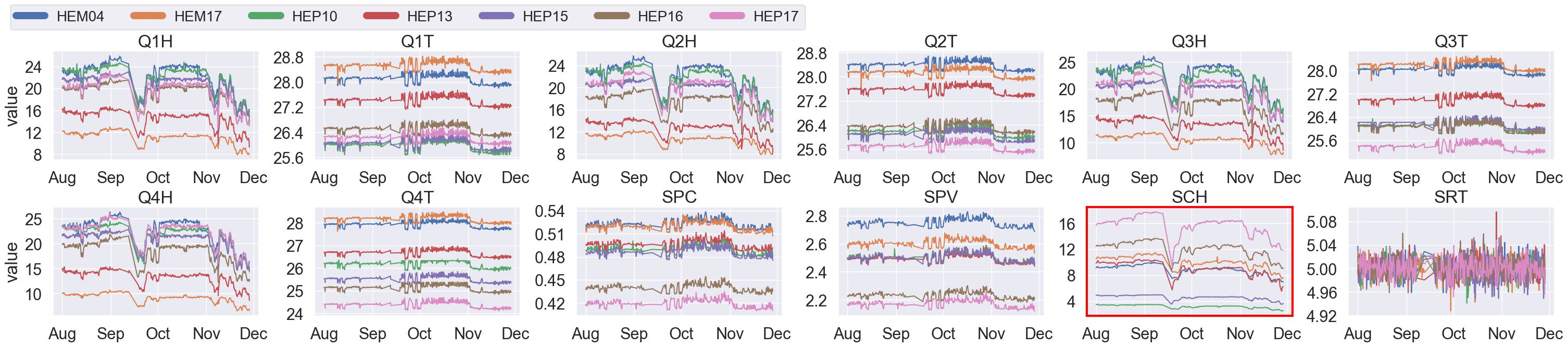}
\end{subfigure}

\begin{subfigure}[]{1.00\textwidth}
\centering
\includegraphics[width=\linewidth]{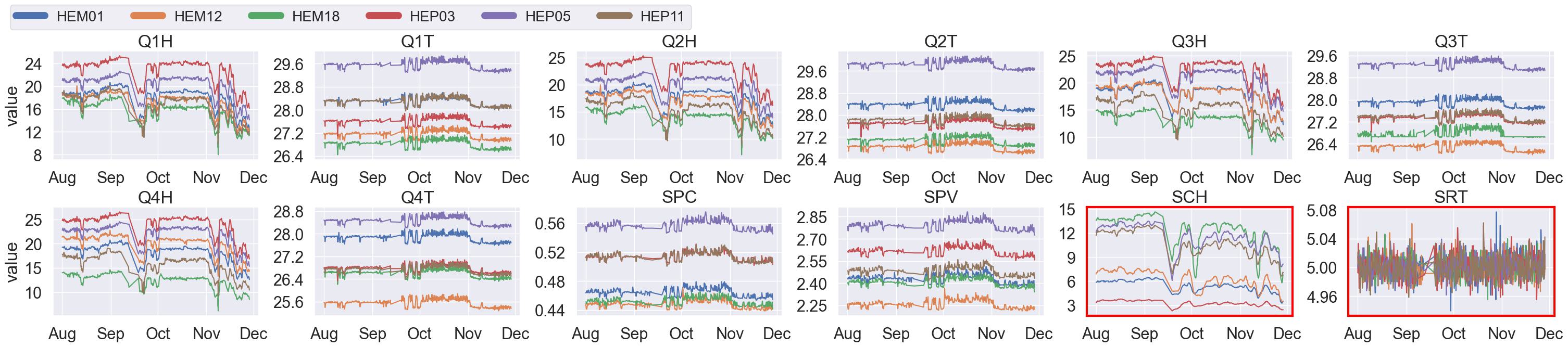}
\end{subfigure}

\begin{subfigure}[]{1.00\textwidth}
\centering
\includegraphics[width=\linewidth]{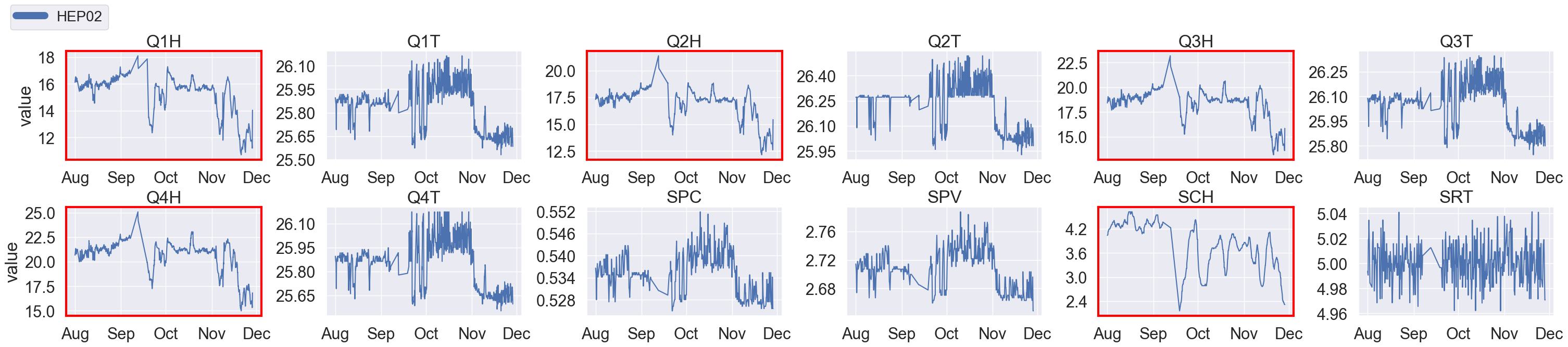}
\end{subfigure}

\caption{Sensor data of the RBX clusters $1,\dots,5$ (top to bottom). Diverging patterns in the SCH across the clusters in October and November; bigger humps on the Q[1-4]H at the beginning of September and smaller jumps on the Q[1-4]T, SPV, and SPC in cluster-1 at the end of September. The root-cause sensors that contributed most to the system clustering divergence are highlighted with a red box.}
\label{fig:rca__ts_signals_clthr}
\end{figure*}

\begin{figure*}[!htbp]
\centering
\begin{subfigure}[]{1.00\textwidth}
\centering
\includegraphics[width=\linewidth]{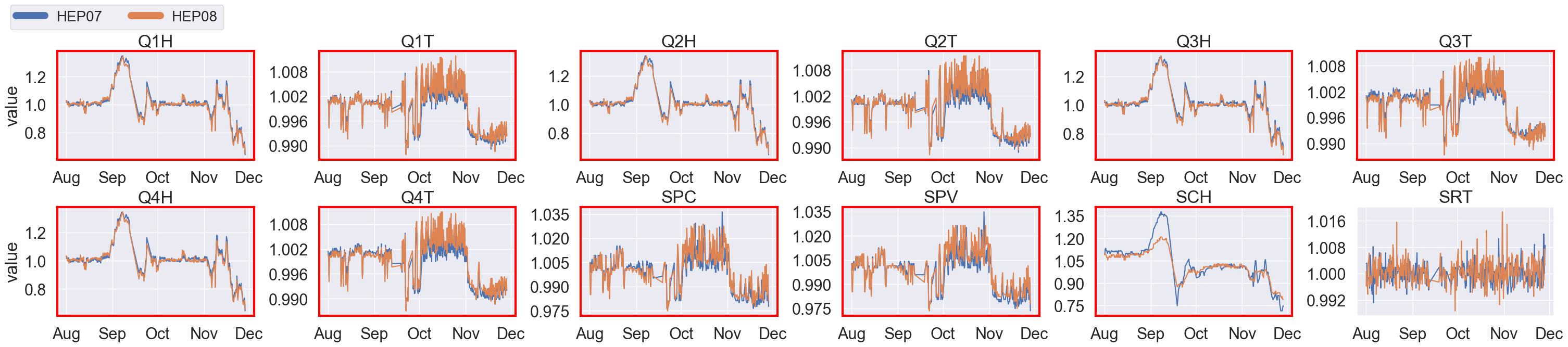}
\end{subfigure}

\begin{subfigure}[]{1.00\textwidth}
\centering
\includegraphics[width=\linewidth]{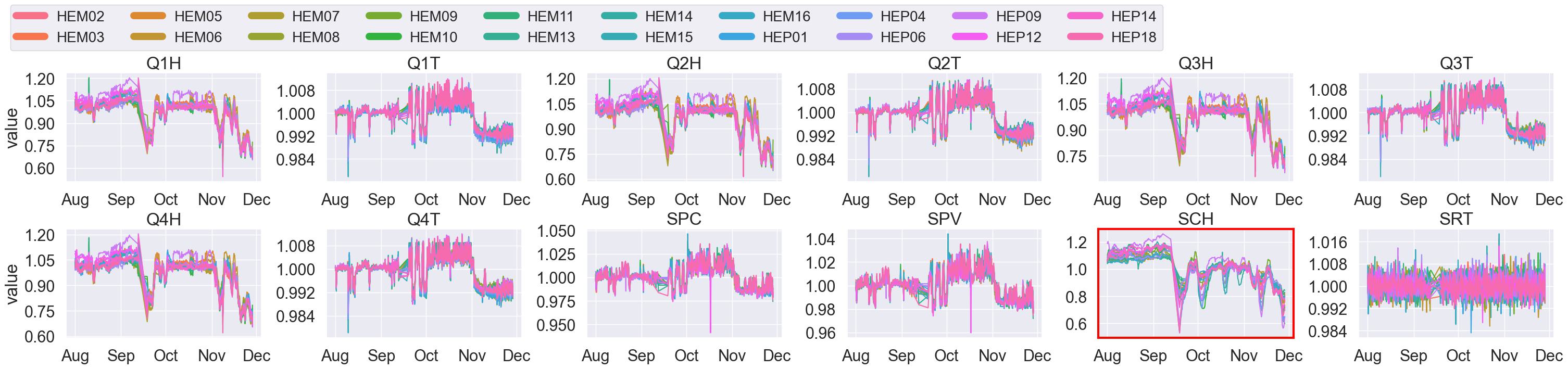}
\end{subfigure}

\begin{subfigure}[]{1.00\textwidth}
\centering
\includegraphics[width=\linewidth]{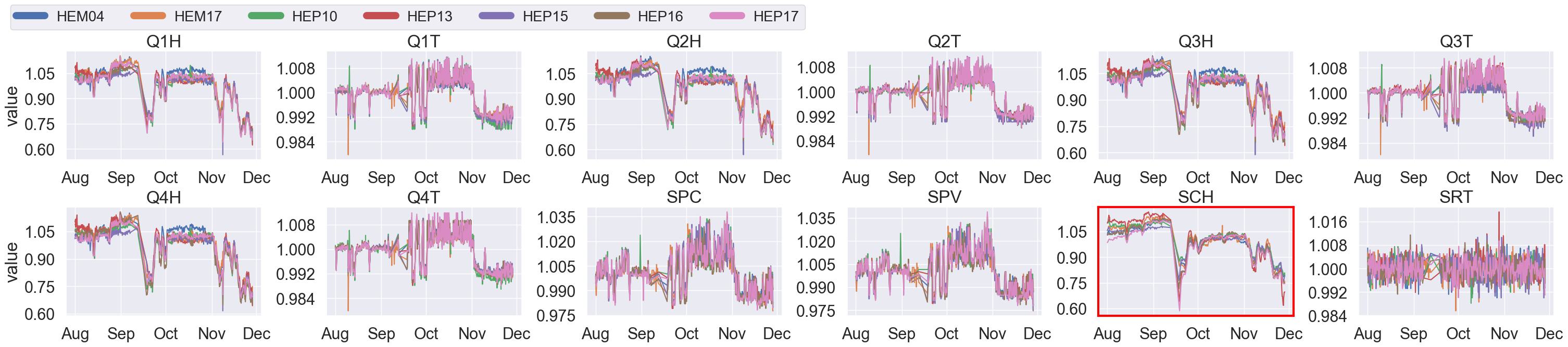}
\end{subfigure}

\begin{subfigure}[]{1.00\textwidth}
\centering
\includegraphics[width=\linewidth]{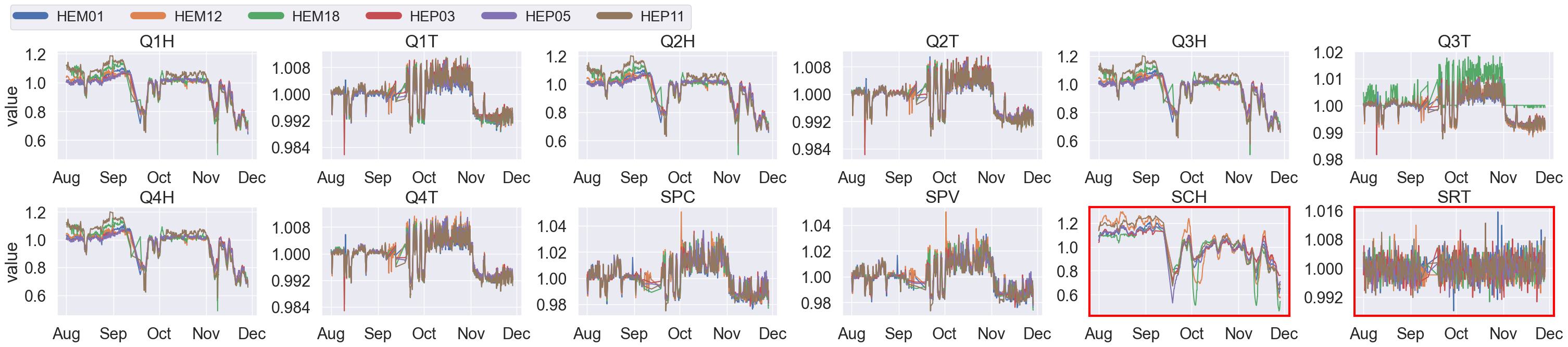}
\end{subfigure}

\begin{subfigure}[]{1.00\textwidth}
\centering
\includegraphics[width=\linewidth]{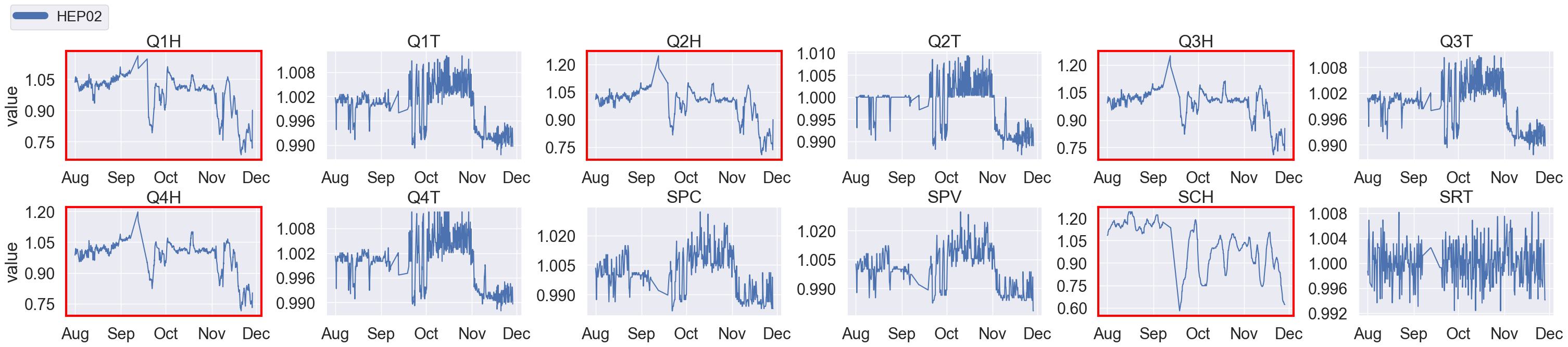}
\end{subfigure}

\caption{Normalized sensor data (divided by median value) of the RBX clusters $1,\dots,5$ for enhanced illustration. Diverging patterns in the SCH across the clusters in October and November; bigger humps on the Q[1-4]H at the beginning of September and smaller jumps on the Q[1-4]T, SPV, and SPC in cluster-1 at the end of September. The root-cause sensors, that contributed most to the system clustering divergence, are highlighted with a red box.}
\label{fig:rca__ts_signals_norm_clthr}
\end{figure*}

Our proposed data-driven interconnection analysis conforms with the expected behavior of the RBX-RM systems. The QIE cards in a given RM are in close proximity, and their environmental characteristics, such as temperature and humidity readings, correlate strongly. The SPV and SPC sensors are from the Peltier system, which controls the temperature of the RM internal systems. External humidity controllers such as nitrogen gas regulate the humidity levels of the RMs. 
The interconnections between the humidity of the SiPM SCH and QIE Q[1-4]H sensors are not strong due to divergence in some RBX systems, i.e., CL-4 and CL-5. The HCAL external humidity controllers regulate a group of RBXes that may cause divergence in humidity measurement among the RBXes.
Figure~\ref{fig:rca__ts_signals_clthr} 
and \ref{fig:rca__ts_signals_norm_clthr} 
displays the time series data 
(raw and normalized, respectively) 
of the sensors grouped by the RBX system clusters, validating the accuracy of the generated knowledge of the interconnection analysis that similar RBXes are grouped, diverging patterns in the SCH sensors across clusters in October and November, and humps in the Q[1-4]H sensors of CL-1 at the beginning of September.

\section{Conclusion}
\label{sec:conclusion}

We have developed a simple online mechanism---without heavy data preprocessing, data annotation, and pre-training requirements---for fast discovery of divergent behaviors among multi-systems through interconnection analysis. Our experiment demonstrates the promising efficacy of our approach in exploring outlying behaviors on the readout systems of the Hadron Calorimeter at the Compact Muon Solenoid experiment. 
Our approach enables real-time and computationally efficient discrepancy discovery for multi-systems with multi-level clustering analysis. The tool broadly applies to condition monitoring of multiple systems with multivariate environments. The real-time processing performance depends on several factors, including the similarity measuring algorithm, the number of systems and sensors, and the data characteristics. Data transformation techniques, such as converting time to frequency or time-frequency domain and data encoding through advanced temporal data embedding, can be beneficial but add computational overhead.

\bibliographystyle{IEEEtran}

\footnotesize
\bibliography{main.bbl}

\end{sloppypar}
\end{document}